\pdfoutput=1

\documentclass[11pt]{article}

\usepackage[]{acl}
\usepackage{booktabs}
\usepackage{multirow}
\usepackage{bbding}
\usepackage{times}
\usepackage{latexsym}
\usepackage[T1]{fontenc}

\usepackage[utf8]{inputenc}

\usepackage{microtype}

\usepackage{inconsolata}
\usepackage{graphicx}
\usepackage{comment}
\usepackage{bm}
\usepackage{makecell}
\usepackage{amsmath}
\usepackage{footmisc}
\usepackage{amssymb}
\usepackage{xcolor}
\usepackage{pifont}
\usepackage{makecell}
\usepackage{colortbl}
\newcolumntype{C}[1]{>{\centering\arraybackslash}p{#1}}

\definecolor{ballblue}{rgb}{0.13, 0.67, 0.8}
\definecolor{bananamania}{rgb}{0.98, 0.91, 0.71}
\definecolor{lazure}{rgb}{0.54, 0.81, 0.94}
\definecolor{azure}{rgb}{0.0, 0.5, 1.0}
\definecolor{dgreen}{rgb}{0.53, 0.66, 0.42}
\definecolor{lgreen}{rgb}{0.64, 0.76, 0.68}
\definecolor{candypink}{rgb}{0.89, 0.44, 0.48}
\definecolor{lorange}{rgb}{0.98, 0.81, 0.69}
\definecolor{dorange}{rgb}{1.0, 0.6, 0.4}
\definecolor{deepgreen}{rgb}{0.0, 0.5, 0.0} 
\title{Let's Learn Step by Step: Enhancing In-Context Learning Ability with Curriculum Learning}

\author{Yinpeng Liu, Jiawei Liu, Xiang Shi, Qikai Cheng, Yong Huang
\and Wei Lu \\
        School of Information Management, Wuhan University, China\\
        Information Retrieval and Knowledge Mining Laboratory, Wuhan University, China\\
        \texttt{\{yinpengliu, laujames2017, coding, chengqikai, yonghuang1991, weilu\}@whu.edu.cn}}

\begin{document}
\maketitle
\begin{abstract}
Demonstration ordering, which is an important strategy for in-context learning (ICL), can significantly affects the performance of large language models (LLMs). However, most of the current approaches of ordering require high computational costs to introduce the priori knowledge. In this paper, inspired by the human learning process, we propose a simple but effective demonstration ordering method for ICL, named the few-shot In-Context Curriculum Learning (ICCL). The ICCL implies gradually increasing the complexity of prompt demonstrations during the inference process. The difficulty can be assessed by human experts or LLMs-driven metrics, such as perplexity. Then we design extensive experiments to discuss the effectiveness of the ICCL at both corpus-level and instance-level. Moreover, we also investigate the formation mechanism of LLM's ICCL capability.
Experimental results demonstrate that ICCL, developed during the instruction-tuning stage, is effective for representative open-source LLMs. To facilitate further research and applications by other scholars, we make the code publicly available\footnote{https://github.com/61peng/curri\_learning}.
\end{abstract}

\section{Introduction}

Human education is methodical and incremental, building upon previously accumulated knowledge, which inspires \textbf{curriculum} based algorithm designs in machine learning. Curriculum learning, introduced by \citet{bengio2009curriculum}, is originally a method that progressively raises the difficulty of the data samples utilized in the training process. Many studies have demonstrated the efficacy of curriculum learning applied in different models~\citep{portelas2020teacher, nagatsuka-etal-2021-pre} and different tasks~\citep{wang-etal-2022-hierarchical, xu-etal-2020-curriculum}.

\begin{figure}[h]
    \centering
    \includegraphics[width=0.48\textwidth]{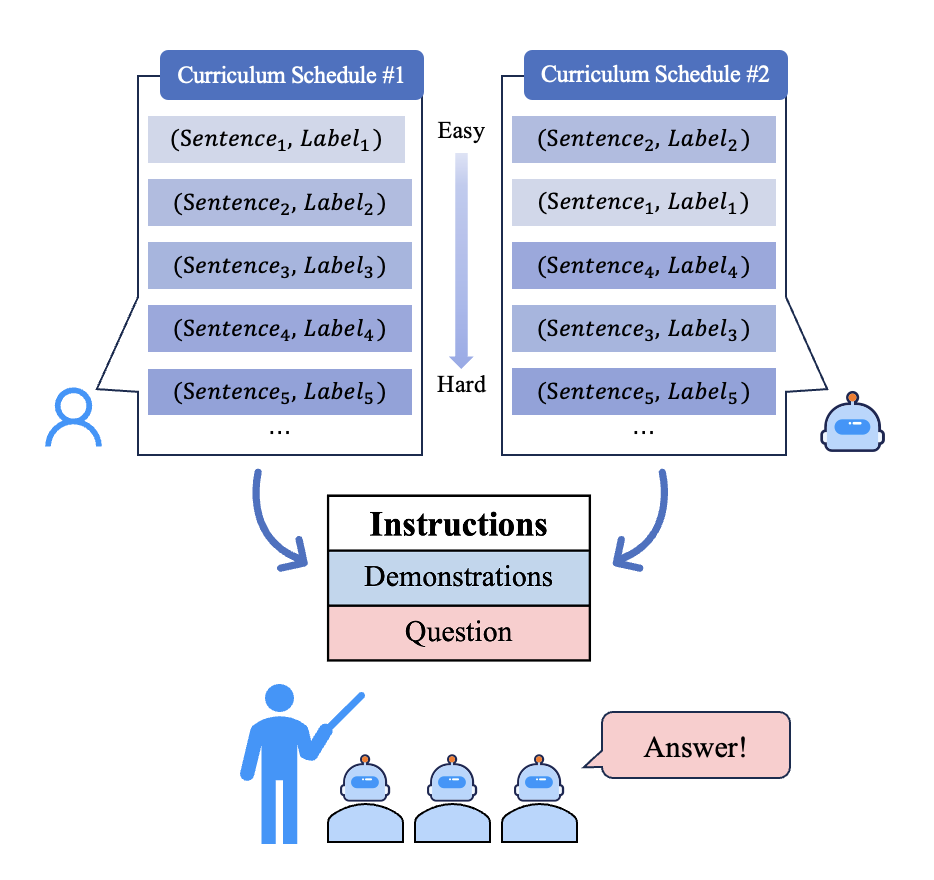}
    \caption{Illustration of In-Context Curriculum Learning (ICCL). The curriculum schedule can be designed by both human and LLMs, schedule constructor sort demonstrations from easy to hard based on their understanding.}
    \label{fig:curr}
\end{figure}

Since instruction-tuned Large Language Models (LLMs) exhibited remarkable proficiency in understanding human intentions and generating human-like text~\citep{ouyang2022training}, researchers have initiated the integration of curriculum learning during instruction tuning~\citep{feng2023citing, lee2023instruction}. 
Aforementioned works demonstrate that curriculum learning facilitates accelerated convergence and the identification of better local minima during the parameter updating process. However, research on the effectiveness of curriculum learning within In-Context Learning (ICL) remains limited.
Some methods that gradually prompt LLMs within instruction, such as Chain of Thought (CoT)~\citep{wei2022chain}, have significantly enhanced the ability of model to perform complex reasoning. This inspires us to apply curriculum learning for ICL.

LLMs with varying performance are treated as students with varying learning abilities, and a human educator plays the role of a facilitator, guiding the learners through the curriculum. Under the human-led curriculum, the models are gradually prompted to solve complex tasks.
Is such a curriculum schedule effective, particularly when compared with many superior demonstration ordering algorithms? If this is the case, at what point is the model's capacity to learn from a curriculum curriculum established? 

To answer those questions, we propose the In-Context Curriculum Learning (ICCL), as illurstrated in Figure~\ref{fig:curr}. The ICCL framework encompasses two roles: curriculum constructor and curriculum learner. The curriculum constructor, which could be either human experts or LLMs, ranks the demonstrations based on their comprehension of difficulty. Subsequently, the learner is guided in progressively solving tasks.


Our main contributions are as follows: (1) We propose the ICCL, a straightforward and effective demonstration ordering method, and validate the effectiveness of ICCL for open-source LLMs. (2) We adopt perplexity as the metric to assess the difficulty of demonstration, which outperformed many superior demonstration ordering methods. (3) Comparative analysis indicates that the ICCL capability of LLMs is developed during the instruction-tuning stage.

\section{Related Work}

\paragraph{Demonstrations Organization}
Numerous studies~\citep{dong2022survey, wan2023efficient} show that the performance of LLMs is heavily influenced by the selection and ordering of demonstrations, indicating that different organizational approaches lead to the assimilation of distinct semantic information.

\citet{lu-etal-2022-fantastically} find that the performance of pretrained language models can vary from nearly state-of-the-art to random guess performance depending on how samples are ordered. This implies there exist multiple strategies for arranging prompt orders to enhance performance. They identify outstanding demonstration organizations based on entropy statistics.
\citet{liu-etal-2022-makes} retrieve demonstrations that are semantically-similar to test source and order them by increasing cosine similarity.
\citet{wu-etal-2023-self} propose a ranking algorithm inspired by the compression viewpoint, which considers the codelength required to compress and transmit testing label. The codelength can be calculated using Shannon-Huffman code.

Since the existing research has substantiated that demonstration organization can significantly affects performance, we aim to delve into the optimization of prompt orders in ICL. To this end, we introduce curriculum learning strategies successfully employed in machine learning into ICL.
\paragraph{Curriculum Learning} 

The concept of curriculum learning~\citep{bengio2009curriculum} has inspired numerous research to address various natural language processing tasks.
\citet{wang-etal-2022-hierarchical} proposes a novel framework for Abstract Meaning Representation (AMR) parsing using hierarchical curriculum learning, achieving significant improvements on AMR2.0 and AMR3.0 benchmarks.
\citet{jia-etal-2023-sample} introduces an approach applying curriculum learning to natural language generation (NLG) tasks. The authors propose a strategy that starts by training models to generate the final few words of a sequence, progressively extending to generate the entire sequence.
However, the aforementioned studies all require adjustments to model parameters. There is currently a lack of research exploring curriculum learning in context.

\section{Methodology} \label{sec:iccl}

Inspired by curriculum learning employed in training process, we investigate a novel few-shot In-Context Curriculum learning (ICCL), which is essentially a strategy for ordering demonstrations: sort the demonstration examples in order of increasing difficulty. This ordering strategy prompt LLM to absorb many skills and tasks within the parameters gradually.

\begin{table*}[tbp]
  \centering
  \footnotesize
  \resizebox{0.82\textwidth}{!}{
    \begin{tabular} {l|ccccc|c}
    \toprule
    \multirow{3}{*}{\textbf{Method}} & \multicolumn{2}{c}{\textbf{\footnotesize{SciCite}}}  & \multicolumn{2}{c}{\textbf{\footnotesize{SciNLI}}}  & \multicolumn{1}{c}{\textbf{\footnotesize{SciERC}}} & \multicolumn{1}{|c}{\textbf{\footnotesize{Overall}}}\\
    \cmidrule(lr){2-3} \cmidrule(lr){4-5} \cmidrule(lr){6-6} \cmidrule(lr){7-7}
     & \textit{Macro P} & \textit{Macro} $F_1$  & \textit{Accuracy} & \textit{Macro} $F_1$ & \textit{Micro} $F_1$ & \textit{Avg} $F_1$ \\ 
    
    \midrule
    \multicolumn{4}{l}{\textsc{Mixtral-8x7B-Instruct-v0.1}} \\
    Random & 
    69.74\textsubscript{$\pm$3.76} & 62.57\textsubscript{$\pm$0.94} & 
    42.38\textsubscript{$\pm$0.11} & 37.21\textsubscript{$\pm$0.10} & 
    23.91\textsubscript{$\pm$0.32} & 41.23\\
    VoteK & 
    \cellcolor{lazure}68.82\textsubscript{$\pm$2.11} & \cellcolor{azure}49.88\textsubscript{$\pm$1.81} & 
    \cellcolor{azure}38.89\textsubscript{$\pm$0.08} & \cellcolor{azure}31.66\textsubscript{$\pm$0.44}  & 
    \cellcolor{orange}30.24\textsubscript{$\pm$1.70}  & \cellcolor{azure}37.26\\
    \textbf{ICCL(Ours)} & 
    \cellcolor{orange}71.32\textsubscript{$\pm$1.58} & \cellcolor{orange}66.76\textsubscript{$\pm$2.65} & 
    \cellcolor{orange}52.21\textsubscript{$\pm$0.28} & \cellcolor{orange}49.87\textsubscript{$\pm$0.26} & 
    \cellcolor{lorange}24.90\textsubscript{$\pm$0.74}  & \cellcolor{orange}47.18\\
    
    \midrule
    
    \multicolumn{4}{l}{\textsc{Llama 2-70B-chat}} \\
    Random & 
    64.99\textsubscript{$\pm$0.45} & 59.37\textsubscript{$\pm$0.18} &
    39.68\textsubscript{$\pm$0.45} & 34.31\textsubscript{$\pm$0.38} &
    24.40\textsubscript{$\pm$5.28} & 39.36\\
    VoteK & 
    \cellcolor{lazure}61.27\textsubscript{$\pm$1.09} & \cellcolor{orange}63.11\textsubscript{$\pm$0.25} & 
    \cellcolor{lazure}37.53\textsubscript{$\pm$0.20} & \cellcolor{lazure}27.46\textsubscript{$\pm$0.20} & 
    \cellcolor{orange}30.54\textsubscript{$\pm$0.60} & \cellcolor{orange}40.37\\
    \textbf{ICCL(Ours)} & 
    \cellcolor{orange}67.58\textsubscript{$\pm$2.84} & \cellcolor{orange}62.56\textsubscript{$\pm$1.28} & 
    \cellcolor{orange}41.13\textsubscript{$\pm$0.39} & \cellcolor{orange}35.59\textsubscript{$\pm$0.52} &
    \cellcolor{orange}31.45\textsubscript{$\pm$0.90}   & \cellcolor{orange}43.20 \\
    \midrule
    
    \multicolumn{4}{l}{\textsc{Qwen1.5-72B-Chat}} \\
    Random & 
    75.19\textsubscript{$\pm$0.75} & 74.70\textsubscript{$\pm$0.38} & 
    48.38\textsubscript{$\pm$0.24} & 45.85\textsubscript{$\pm$0.31} & 
    19.51\textsubscript{$\pm$0.32}  & 46.59\\
    VoteK & 
    \cellcolor{orange}77.82\textsubscript{$\pm$0.22} & \cellcolor{lorange}75.62\textsubscript{$\pm$0.22} & 
    \cellcolor{orange}49.88\textsubscript{$\pm$0.40} & \cellcolor{orange}47.55\textsubscript{$\pm$0.51} & 
    \cellcolor{orange}30.19\textsubscript{$\pm$0.63}  & \cellcolor{orange}51.12\\
    \textbf{ICCL(Ours)} & 
    \cellcolor{orange}76.98\textsubscript{$\pm$0.45} & \cellcolor{lorange}75.02\textsubscript{$\pm$0.87} & 
    \cellcolor{orange}50.83\textsubscript{$\pm$0.31} & \cellcolor{orange}49.09\textsubscript{$\pm$0.30} & 
    \cellcolor{orange}26.68\textsubscript{$\pm$0.15}   & \cellcolor{orange}50.26\\
    
    \midrule
  \end{tabular}
  }
  \caption{Evaluation result of mainstream LLMs applying ICCL on three scientific datasets at corpus level. We adopt $F_1$ score as the core metric and perform averaging to get overall $F_1$ score. All the results are calculated based on 3 different random seeds over test set of each task. standard deviation are in small font. Dark and light blue colored cells stand for decline $> 1\%$ and $< 1\%$ compared to \textbf{Random} baseline, respectively. Dark and light orange colored cells stand for improvement $ > 1\%$ and $ < 1\%$, respectively.} 
  \label{tab:one}
\end{table*}

\subsection{Problem Formulation}

Given a LLM $\theta$, there are $n$ demonstrations $\{(x_i, y_i)\}_{i=0}^n$ selected to instruct $\theta$ to solve specific task $\mathcal{T}$. 
While trying to adapt $\theta$ for $\mathcal{T}$, different demonstration orders $\mathcal{D}$ have different efficiency in utilizing parameters $\theta$, the parameter-efficiency $E_p$ is measured by performance metrics. 
We hypothesizes that when demonstrations are arranged from simple to difficult, it will increase the model's $E_p$ ad much as possible.
Therefore, the objective of ICCL is to acquire an order $\mathcal{D}_{\text{curriculum}}$ that:

\begin{equation}
D_{\text{curriculum}} \approx \underset{D}{{\arg\max}} \, E_p(\{(x_i, y_i)\}_{i=0}^n; \theta)
\end{equation}

ICCL remain parameters $\theta$ fixed throughout the process and merely modifies the ordering of $\mathcal{D}$ to progress from simple to complex. Consequently, the crux lies in the method of measuring the complexity of demonstrations.

\subsection{Curriculum Schedule Construction}

We firstly rely on human experts to construct a curriculum-based context for LLMs at corpus level. 
Specifically, we engaged five human experts, ranging from undergraduates to professors, to rank the demonstrations based on their perceived difficulty,  The final ordering was determined by averaging the rankings provided by each expert. To ensure the reliability of the final order, we employed Kendall's coefficient of concordance as the agreement scores among experts. At instance level, appropriate samples can be selected for each test target using demonstrations retrieval algorithms (such as TopK~\citep{liu-etal-2022-makes}). The ranker shifts from humans to LLMs. While human experts can judge the demonstrations difficulty based on their understanding, LLMs may not perceive it the same way. An intuitive approach is to use \textbf{perplexity} to quantify the LLMs' understanding of complexity.

We retrieve \(n\) candidate samples that are most similar to the test target. Then, we calculate the complexity of each sample using:
\begin{equation}
    \text{Comp}(x_i, y_i) = \exp\left\{- \log p(y_i | \mathcal{I_\theta}(x_i))\right\}
\end{equation}
where \((x_i, y_i)\) represents a sample in the candidate set, and \(\mathcal{I_\theta}(x_i)\) denotes the instruction template of LLM $\theta$ with input \(x_i\). We measure the complexity of a sample by calculating the perplexity on the label \(y_i\) given the specified instruction, and order the demonstrations $\{(x_i, y_i)\}$  with lower perplexity first.

\section{Experiments}

\subsection{Setup} 


\paragraph{Datasets}
Scientific papers are relatively complex discourse in the structured educational journey of human. The comprehension of scientiﬁc text requires domain expertise and logical reasoning capability. Therefore, corpus composed of scientific texts is selected as the benchmark evaluation set for estimating the effectiveness of curriculum learning applied during the inference stage of LLMs.

We evaluate ICCL on three scientific dataset: SciCite~\citep{cohan-etal-2019-structural}, SciNLI~\citep{sadat-caragea-2022-scinli} and SciERC~\citep{luan-etal-2018-multi}, encompassing tasks in text classification, natural language inference and information extraction.

\paragraph{Models}We utilize a range of open-source performant LLMs, including LlaMA2~\citep{touvron2023llama}, Mixtral-8x7B~\citep{jiang2024mixtral}, Qwen1.5~\citep{bai2023qwen}, exploring their applications within the ICCL framework.

\paragraph{Baseline}
For corpus-level methods, we consider \textbf{VoteK}~\citep{su2022selective} and a \textbf{Random} baseline. 
For instance-level methods, we select \textbf{KATE}~\citep{liu-etal-2022-makes}, \textbf{TopK+LocalE}~\citep{lu-etal-2022-fantastically}, \textbf{TopK+MDL}~\citep{wu-etal-2023-self} and a \textbf{TopK} baseline that select 5 demonstrations that are semantically closest to testing samples and rank them randomly.

More experiment details are present in Appendix.

\begin{table}[!t]
    \centering
    \resizebox{0.48\textwidth}{!}{
    \begin{tabular}{llccc}
    \toprule
        \textbf{LLM} & \textbf{Method} & \textbf{SciCite} & \textbf{SciNLI} & \textbf{SciERC} \\ 
        \midrule
         \multirow{5}{*}{\makecell[l]{Mixtral-8x7B\\-Instruct-v0.1}} & TopK & 64.85 & 34.69 & 32.98     \\
        &~~+~KATE & 64.19 & 35.20 & 32.48      \\ 
        & ~~+~LocalE & 67.13 & 33.16 & - \\ 
        &~~+~MDL & 67.06 & 35.22 &  -  \\ 
        & \cellcolor{bananamania}~~+~\textbf{ICCL(Ours)} & \cellcolor{bananamania}\textbf{67.92} & \cellcolor{bananamania}\textbf{37.58} & \cellcolor{bananamania}\textbf{33.58}   \\ 
        \midrule
         \multirow{5}{*}{\makecell[l]{Llama2\\-70B-Chat}} & TopK & 60.86 & 39.70 & 38.54     \\
        &~~+~KATE & 57.11 & 39.81 & 38.71      \\ 
        & ~~+~LocalE & 59.30 & 39.95 & - \\ 
        &~~+~MDL & \textbf{64.11} & 40.02 &  -  \\ 
        & \cellcolor{bananamania}~~+~\textbf{ICCL(Ours)} & \cellcolor{bananamania}63.30 & \cellcolor{bananamania}\textbf{40.48} & \cellcolor{bananamania}\textbf{39.03}   \\ 
    \bottomrule
    \end{tabular}
    }
    \caption{Evaluation result of LLMs applying Instance-Level ICCL on three scientific datasets. Numbers in bold indicate the highest $F_1$ among all methods. All the results are calculated based on 3 different random seeds over test set of each task.} 
    \label{tab:two}
\end{table}

\subsection{Main Result}

A comparison between 3 mainstream open-source LLMs across 3 NLP tasks shows the superiority of ICCL over other corpus-level methods, depicted in Table~\ref{tab:one}. 
The demonstrations for both ICCL and the random baseline were selected by human experts, while VoteK, built upon TopK, selected diverse yet representative examples using voting mechanism. ICCL maintains an ordering from simple to complex for all test samples, whereas the random and VoteK baseline employs a random order. 

Despite VoteK achieving commendable performance in certain settings, it exhibits a large standard deviation, indicating instability in its improvements. For Mixtral-8x7B, the performance of VoteK is approximately 10\% lower than random baseline. While ICCL consistently achieves stable improvements across all LLMs and NLP tasks compared to random baseline, and shows an average improvement of 9\% compared to random baseline, and 7.8\% compared to VoteK. This demonstrates that \textbf{heuristic curriculum learning methods are effective for in-context learning}. 




At instance level, we employ the topK algorithm for demonstrations retrieval to construct candidate set. Subsequently, we utilize ordering algorithms such as KATE, LocalE, and MDL to generate performant prompts from this candidate set as baselines. As shown in Table~\ref{tab:two}, ICCL still registers decent improvements on most tasks compared with instance-level baselines. Our method achieves an overall improvement of 4.96\% compared to TopK with random order and 5.48\% compared to KATE for Mixtral. We also notice that LocalE and MDL are competitive on SciCite. However, these two methods are only suitable for classification tasks with a limited search space and are difficult to apply to complex generative tasks. \textbf{This highlights the versatility of our method, indicating that it can be applied to a wide range of NLP tasks}.

\begin{figure}[!t]
    \centering
    \includegraphics[width=0.48\textwidth]{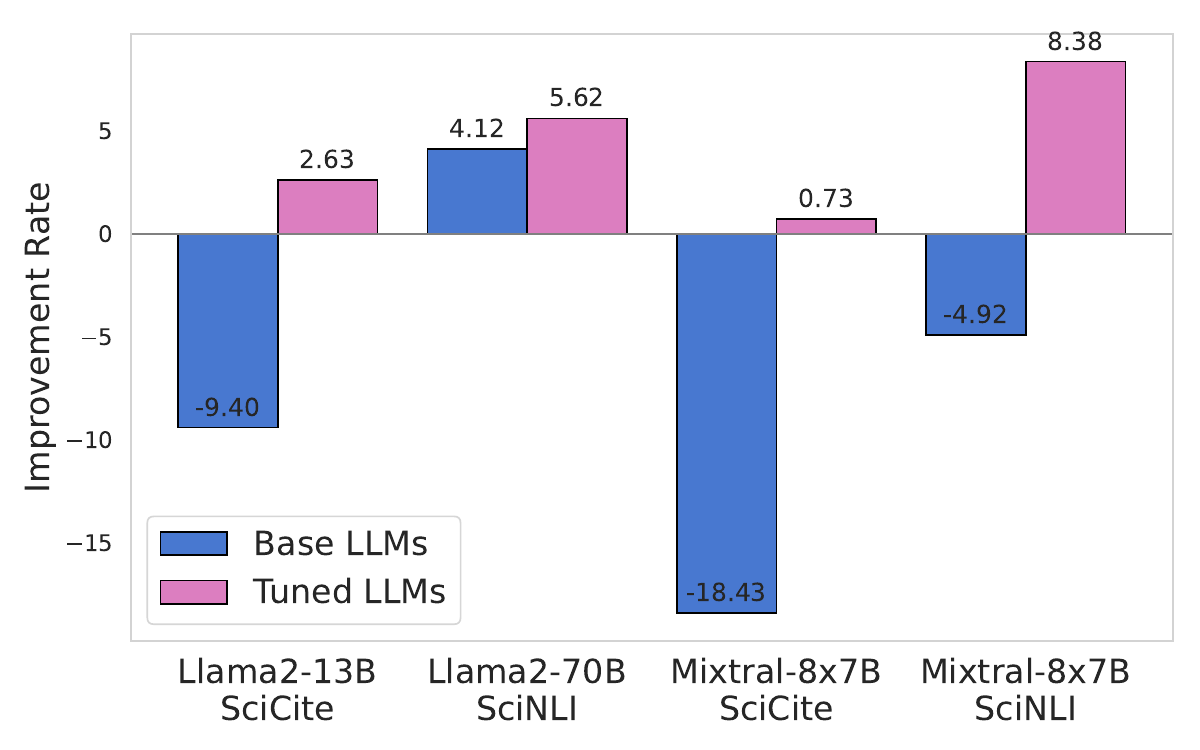}
    \caption{$F_1$ scores improvement (or decline) rates for both Based LLMs and Instruction-Tuned LLMs using ICCL compared with random order.}
    \label{fig:bar_curr}
\end{figure}

\subsection{Formation Mechanism of ICCL Capability}
To explore the formation mechanism of ICCL, i.e., whether the model's ability to learn the curriculum from context is established during pre-training or instruction-tuning, we conduct ICCL experiments on both the base models and the instruction-tuned models separately. 

Figure~\ref{fig:bar_curr} shows that the performance of ICCL on the base model is unstable, with an average decrement of 7.16\%. Even in instances where an improvement is observed, it is weaker compared to the enhancement effect on the corresponding instruction-tuned models. This evidences the base models' lack of sensitivity to the curriculum-based demonstration order. It further intimates that \textbf{the competency for ICCL is most likely acquired during the instruction-tuning stage}.

\section{Conclusion}
In this work, we argue that gradually increase the complexity of the demonstrations in prompt can achieve better performance. To substantiate this claim, we propose In-Context Curriculum Learning (ICCL), a straightforward yet effective demonstration ordering method for both corpus-level and instance-level. 
We design three sets of experiments to exploring the validity and mechanism of curriculum learning within context. The experimental results affirm the effectiveness of ICCL on open-source LLMs.

\section*{Limitations}
Due to the limited timeframe of the experiment, we were unable to utilize the latest LLMs, such as Meta Llama 3~\citep{llama3modelcard}. However, our experiments with the current mainstream models have demonstrated the effectiveness of the heuristic method of in-context curriculum learning. In future research, we will incorporate more recent LLMs to ensure the robustness of our method across different models.



\bibliography{ref}

\appendix
\section{Datasets}
Dataset information is detailed in Table~\ref{tab:dataset}.
Here are examples of each datasets:
\begin{itemize}
    \item \textbf{SciCite}
    \begin{itemize}
        \item[--] \textit{Sentence:} A direct consequence is that overheads for address translation have grown to dominate run time for many important workloads with large memory footprint [46, 113, 241, 302, 303, 375].
        \item[--] \textit{Label:} background
    \end{itemize}
    \item \textbf{SciERC}
    \begin{itemize}
        \item[--] \textit{Sentence:} This paper presents an approach to the unsupervised learning of parts of speech which uses both morphological and syntactic information.
        \item[--] \textit{Label:} [['approach', 'Generic'], ['unsupervised learning of parts of speech', 'Task'], ['morphological and syntactic information', 'OtherScientificTerm']]
    \end{itemize}
    \item \textbf{SciNLI}
    \begin{itemize}
        \item[--] \textit{Sentence1:} Without L alignment , we observe a reduction in both accuracy and BLEU on Yelp.
        \item[--] \textit{Sentence2:} this tendency is inconsistent on Amazon (i.e., -2.2 accuracy and +0.56 BLEU).
        \item[--] \textit{Label:} contrasting
    \end{itemize}
\end{itemize}

\begin{table}[]
    \centering
    \resizebox{\linewidth}{!}{
    \begin{tabular}{l|cccc}
    \toprule
    \textbf{Dataset} & \textbf{Task} & \textbf{\# Test} & \textbf{\# Labels} & \textbf{\# Demos}\\ \midrule
     \textbf{SciCite}  &  Citation Intent Classification & 1861 & 3 & 5 \\
     \textbf{SciERC}  &  Scientific Information Recognition & 551 & 7 & 5 \\
     \textbf{SciNLI}   &  Scientific Language Inference & 4000 & 4 & 4 \\
    \bottomrule
    \end{tabular}
    }
    \caption{Details of datasets.}
    \label{tab:dataset}
\end{table}

\begin{table*}[h]
    \centering
        \begin{tabular}{C{1.4cm}|p{13.5cm}}
            \toprule
            LLM & \multicolumn{1}{c}{In-context Curriculum Learning Prompts} \\
            \midrule
            Mixtral & \begin{minipage}{13.5cm}
                    \textbf{Template:} [INST]\textbf{\{Task Description\}} + \textbf{\{Sentence\}}[/INST]\textbf{\{Label\}}</s> \\
                    \phantom{\textbf{Template:}} [INST]\textbf{\{Test Input\}}[/INST] \\ 
                    \textbf{Task Description: }You are a scientific literature analyst. Extract scientific entities from sentences. The scientific entity category includes ['Method', 'Task', 'Metric', 'Material', 'Generic', 'OtherScientificTerm', 'Generic'].\\
                    \textbf{Demonstration:} Sentence: \textit{text} $x_i$\\
                    \phantom{\textbf{Demonstration:}} Label: \textit{label} $y_i$ \\
                    \textbf{Test Input:} Sentence: \textit{text} $x$
                \end{minipage} \\
            \midrule
            Llama2 & \begin{minipage}{13.5cm}
                    \textbf{Template:} <s>[INST] <<SYS>> \textbf{\{System Message\}}<</SYS>>\\
                    \phantom{\textbf{Template:}} \textbf{\{Task Description\}} + \textbf{\{Sentence\}}[/INST]\textbf{\{Label\}}</s> \\
                    \phantom{\textbf{Template:}} <s>[INST]\textbf{\{Test Input\}}[/INST] \\ 
                    \textbf{Task Description: }Identify the intent of a citation in scientific papers. Choose the citation intention of the following sentence from ['method', 'background', 'result'].\\
                    \textbf{Demonstration:} Sentence: \textit{text} $x_i$\\
                    \phantom{\textbf{Demonstration:}} Label: \textit{label} $y_i$ \\
                    \textbf{Test Input:} Sentence: \textit{text} $x$
                \end{minipage} \\
            \midrule
            Qwen1.5 & \begin{minipage}{13.5cm}
                    \textbf{Template:} <|im\_start|>system \textbf{\{System Message\}} <|im\_end|> \\
                    \phantom{\textbf{Template:}} <|im\_start|>user\textbf{\{Task Description\}} + \textbf{\{Sentence\}}<|im\_end|> \\
                    \phantom{\textbf{Template:}} <|im\_start|>assistant \textbf{\{Label\}}<|im\_end|> \\
                    \phantom{\textbf{Template:}} <|im\_start|>user\textbf{\{Test Input\}}<|im\_end|> \\ 
                    \textbf{Task Description: }Identify the semantic relationship between the following pair of sentences. The semantic relationship includes ['reasoning', 'entailment', 'contrasting', 'neutral'].\\
                    \textbf{Demonstration:} Sentence1: \textit{text} $x_{i_1}$\\
                    \phantom{\textbf{Demonstration:}} Sentence2: \textit{text} $x_{i_2}$\\
                    \phantom{\textbf{Demonstration:}} Label: \textit{label} $y_i$ \\
                    \textbf{Test Input:} Sentence1: \textit{text} $x_1$ \\
                    \phantom{\textbf{Test Input:}} Sentence2: \textit{text} $x_2$
                \end{minipage}  \\
            \bottomrule
        \end{tabular}
        \caption{The prompt template of In-Context Curriculum Learning for each LLM.}
        \label{tab:prompt}
\end{table*}

\section{Model}
\label{sec:model-detail}

We use both publicly available and proprietary LLMs with the different model size as follow:
\begin{itemize}
    \item \textbf{LlaMA2}\citep{touvron2023llama} is a collection of pretrained and fine-tuned LLMs developed by MetaAI. For the base model, we select \textit{Llama 2-13B} and \textit{Llama 2-70B}, while for the tuned model, we select \textit{Llama 2-13B-Chat} and \textit{Llama 2-70B-Chat}.
    \item \textbf{Mixtral-8x7B}\citep{jiang2024mixtral} is a pretrained generative Sparse Mixture of Experts. We select \textit{Mixtral-8x7B-v0.1} as base model and \textit{Mixtral-8x7B-Instruct-v0.1} as tuned model to be test with ICCL.
    \item \textbf{Qwen1.5}\citep{bai2023qwen} is the LLM family built by Alibaba Cloud. We select \textit{Qwen1.5-72B-Chat}, which is the largest version of Qwen series.
\end{itemize}

The prompt template of each LLM is shown in Table~\ref{tab:prompt}.
\section{Evaluation Detail} \label{sec:eval-detail}


\begin{table*}[h]
    \centering
    \footnotesize
    \begin{tabular}{p{0.8cm}C{6.85cm}C{6.85cm}}
        \toprule
         & \textbf{Manual Select} & \textbf{VoteK} \\
         
        
        \midrule
        Task & \multicolumn{2}{c}{\textbf{Citation Intent Classification}} \\
        \cmidrule(lr){2-3}
        $W$ & 0.968 & -  \\
        \cmidrule(lr){2-3}
        Demos & \begin{minipage}{6.85cm}\textbf{Sentence:} This result is consistent with the conclusions of the aforementioned recent study of (34) and reinforces them from a significantly broader perspective. \textbf{\textcolor{deepgreen}{Label: result}}\\\textbf{Sentence:} To determine the cell velocity, Darcy's law may be used as the constitutive assumption[21],[18],[22],[13]. \textbf{\textcolor{deepgreen}{Label: method}}\\\textbf{Sentence:} This is clearly in contrast to the results of earlier investigations (Laprise\&Peltier1989a, Pierre-humbert\&Wyman1985, Clark\&Peltier1977), where it was found that the criteria for static and dynamic instabilities are simultaneously satisfied. \textbf{\textcolor{deepgreen}{Label: result}}\\\textbf{Sentence:} nest burrows in close proximity of one another appears to be well founded as previously shown by several studies that measured distances between kin vs. non-kin nest burrows, including in long-term data sets(King 1989b; Viblanc et al. 2010; Arnaud, Dobson \& Murie 2012; Dobson et al. 2012). (5) \textbf{\textcolor{deepgreen}{Label: background}}\\\textbf{Sentence:} We employed three modelling approaches that have successfully been applied in previous studies on species distribution (Guisan and Zimmermann 2000): generalised linear models (GLM, i.e. logistic regression in this case), generalised additive models (GAM), and classification and regression trees. \textbf{\textcolor{deepgreen}{Label: method}} \end{minipage}& \begin{minipage}{6.85cm}\textbf{Sentence:} Genotyping of the SNPs was performed as described previously (Gutknecht et al. 2007; Mo\u0308ssner et al. 2006a, b). \textbf{\textcolor{red}{Label: method}}\\\textbf{Sentence:} 1991) and empirical studies (e.g. Kruuk 1978; Kruuk and Parish 1982; Mills 1989; Mills and Gorman 1997; Geffen et al. 1992; Patterson and Messier 2001; Valenzuela and Macdonald 2002). \textbf{\textcolor{red}{Label: background}}\\\textbf{Sentence:} Furthermore, the hospital anxiety and depression scale (HAD) (Snaith and Zigmond, 1986) and the IBS special scale for quality of life (QOL) (IBSQOL) (Drossman et al., 2000) were used. \textbf{\textcolor{red}{Label: method}}\\\textbf{Sentence:} (Massie and Holland, 1984; Ciaramella and Poli, 2001; Uchitomi et al., 2003; Katz et al., 2004; Ell et al., 2005; Boyd et al., 2012; Kim et al., 2012; Mitchell et al., 2012; Palmer et al., 2012; Pirl et al., 2012; Tada et al., 2012; Warmenhoven et al., 2012; Yu et al., 2012). \textbf{\textcolor{red}{Label: background}}\\\textbf{Sentence:}were retrospective (Okada et al., 2002; Tsujimura et al., 2002; Ramasamy et al., 2005) and the two remaining studies were pseudo-randomized controlled studies (Colpi et al., 2009; Ghalayini et al., 2011); these last two studies were randomized based on the waiting list for the operative theatre. \textbf{\textcolor{red}{Label: background}}\end{minipage}\\
        \midrule
        Task & \multicolumn{2}{c}{\textbf{Scientific Information Recognition}} \\
        \cmidrule(lr){2-3}
        $W$ & 0.936 & -  \\
        \cmidrule(lr){2-3}
        Demos & \begin{minipage}{6.85cm}\textbf{Sentence:} Some of our proof techniques are non-standard and may be of independent interest.\textbf{\textcolor{deepgreen}{Label:[]}}\\\textbf{Sentence:}We provide frequentist and Bayesian analyses for this situation. \textbf{\textcolor{deepgreen}{Label: [['Bayesian analyses', 'Method']]}}\\\textbf{Sentence:}This method requires a source-language dependency parser, target language word segmentation and an unsupervised word alignment component. \textbf{\textcolor{deepgreen}{Label: [['method', 'Generic'], ['source-language dependency parser', 'Method'], ['target language word segmentation', 'Method'], ['unsupervised word alignment component', 'Method']]}}\\\textbf{Sentence:} We evaluate our approach against the state-of-the-art techniques and show that our work improves both the quality and the efficiency of entity summarization. \textbf{\textcolor{deepgreen}{Label: [['approach', 'Generic'], ['state-of-the-art techniques', 'Generic'], ['quality', 'Metric'], ['efficiency', 'Metric'], ['entity summarization', 'Task']]}}\\\textbf{Sentence:} The development of such a model appears to be important in several respects: as a device to represent and to use different dialog schemata proposed in empirical conversation analysis; as a device to represent and to use models of verbal interaction; as a device combining knowledge about dialog schemata and about verbal interaction with knowledge about task-oriented and goal-directed dialogs. \textbf{\textcolor{deepgreen}{Label: [['model', 'Generic'], ['device', 'Generic'], ['dialog schemata', 'OtherScientificTerm'], ['conversation analysis', 'Method'], ['device', 'Generic'], ['models', 'Generic'], ['verbal interaction', 'OtherScientificTerm'], ['device', 'Generic'], ['dialog schemata', 'OtherScientificTerm'], ['verbal interaction', 'OtherScientificTerm'], ['task-oriented and goal-directed dialogs', 'Material']]}}\end{minipage}
        & \begin{minipage}{6.85cm}\textbf{Sentence:} The results of a practical evaluation of this method on a wide coverage English grammar are given. \\\textbf{\textcolor{red}{Label: [['method', 'Generic'], ['wide coverage English grammar', 'Method']]}}\\\\\textbf{Sentence:} Second, we show in this paper how a lexical hierarchy is used in predicting new linguistic concepts. \\\textbf{\textcolor{red}{Label: [['lexical hierarchy', 'OtherScientificTerm'], ['linguistic concepts', 'OtherScientificTerm']]}}\\\\\textbf{Sentence:} -LRB- 2 -RRB- Rather than learning from only labelled data, the abundant unlabelled data are exploited. \\\textbf{\textcolor{red}{Label: [['labelled data', 'Generic'], ['abundant unlabelled data', 'Material']]}}\\\\\textbf{Sentence:} (It is argued that the resulting algorithm is both efficient and flexible and is, therefore, a good choice for the parser used in a natural language interface. \\\textbf{\textcolor{red}{Label: ['algorithm', 'Generic'], ['parser', 'Method'], ['natural language interface', 'Task']]}}\\\\\textbf{Sentence:}We will show the experimental results for two corpora and compare them with the results by the NTHU 's statistic-based system, the only system that we know has attacked the same problem. \\\textbf{\textcolor{red}{Label: [\"NTHU 's statistic-based system\", 'Method'], ['system', 'Generic']]}}\end{minipage}\\
        \bottomrule
    \end{tabular}
    \caption{Examples of Demonstrations Selection by Corpus-level Method, where $W$ is Kendall's coefficient of concordance.}
    \label{tab:case}
\end{table*}

At Corpus level, we arbitrarily select 4-5 demonstration examples from training set, which remain consistent across all test samples to eliminate the influence of demonstration selection. Table~\ref{tab:case} show some example of demonstrations selected by different methods. For each ordering setting, the sequence of demonstrations in the prompt for all models is identical. 

To equitably evaluate the curriculum learning capabilities of diverse LLMs, we adopt the test set and evaluation metric inherent to each dataset. Concretely, we report macro-averaged values for SciCite,  micro-averaged values for SciERC, and accuracy and macro $F_1$ for SciNLI.
We select $F_1$ score as the core metric for all tasks to ascertain the efficacy of ICCL. 

To guarantee the reproducibility of our experimental results, We run every experiment with 3 different random seed, and calculate average metric to report in paper.

\end{document}